\documentclass[a4,journal, onecolumn]{IEEEtran}
\usepackage{amsmath,amsfonts}
\usepackage{algorithmic}
\usepackage{algorithm}
\usepackage{array}
\usepackage{textcomp}
\usepackage{stfloats}
\usepackage{url}
\usepackage{verbatim}
\usepackage{graphicx}
\usepackage{cite}
\usepackage{multicol}
\usepackage{xcolor}
\usepackage{algorithm}
\usepackage{pdflscape}
\usepackage{mathtools, cuted}
\usepackage{subcaption}
\usepackage{booktabs}

\begin{document}

\title{All you need is SAMPAT}
%FROM SAMPAT1.TEX. MOST EXAMPLES MOVED TO SUPPLEMENTARY FILES.
\author{Jayadeva and Madhur Aswani\\
Department of Electrical Engineering, Indian Institute of Technology, Delhi}
        % <-this % stops a space

\markboth{Patents Pending}
{Shell \MakeLowercase{\textit{et al.}}: A Sample Article Using IEEEtran.cls for IEEE Journals}

\IEEEpubid{0000--0000/00\$00.00~\copyright~2021 IEEE}

\maketitle

\begin{abstract}
The current state of the art in AI/ML rests on deep neural architectures, which, in general, suffer from a lack of interpretability. Interpretability is crucial to gleaning insights while analyzing experimental data, where quantitative predictions may not be adequate for a scientist. We present a three layer neural architecture, SAMPAT (Smooth Approximation via Multivariate Polynomials and Analytic Transformations), that can provably learn a continuous, everywhere differentiable function, that can approximate any smooth function arbitrarily closely. SAMPAT's approximant can be expressed as a closed and compact algebraic, analytic expression, providing complete interpretability. Experiments on synthetic and benchmark datasets indicate that SAMPAT yields competitive performance with simpler representations. For many tasks, a two layer SAMPAT suffices. By imposing restrictions on the connectivity between neurons, SAMPAT may be used to provide a range of approximants, including regular and trigonometric polynomials, rational expressions, Gaussians, mixtures of Gaussians, as well as arbitrary combinations of the same; without restrictions, it learns a suitable structure. SAMPAT may be used to factorize polynomials and model nonlinear systems. With the addition of skip connections, a 4 to 6 layer SAMPAT is adequate to represent a substantive range of methods widely used in AI/ML, allowing the choice of a model's family, not just its parameters, to also be optimized as part of the learning process.

% We also introduce RAMA (Refinement through Adaptive Minimization of Ambients) as a new way to build complex transformations using multi-layered small models that remain interpretable. RAMA simplifies adaptation of concepts without having to relearn the entire network.
\end{abstract}


\section{Introduction}\label{intro}
\begin{figure}[ht]
    \centering
    \includegraphics[width=0.25\linewidth]{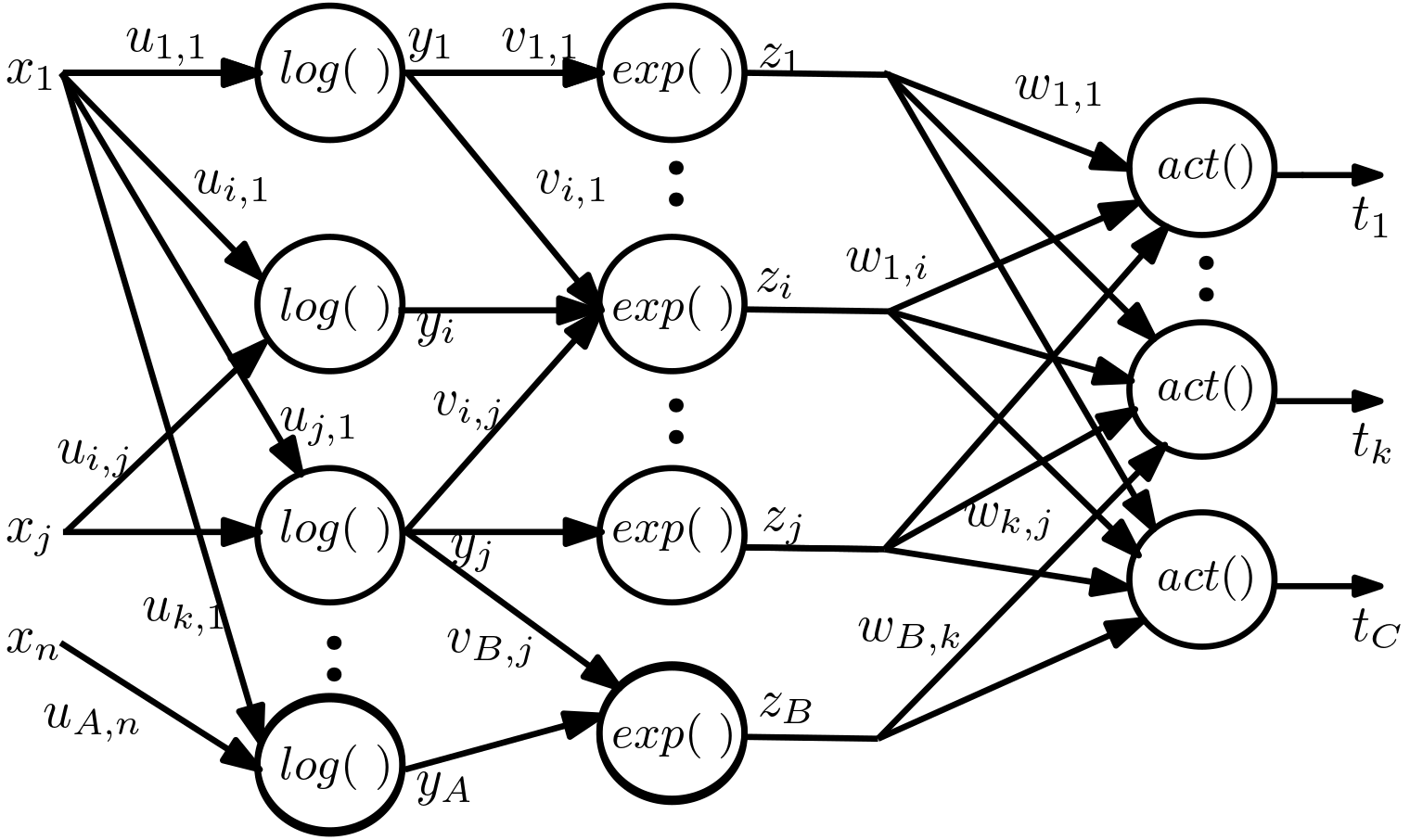}
    \caption{Basic SAMPAT architecture}
    \label{fig:arch1}
    \end{figure}

    It is widely known that a three layer neural network can learn any smooth input-output map. Much work has been done on existence proofs \cite{cybenko1989approximation}. Most of these relate to functions on the unit $n$-dimensional cube. However, constructing a three layer network for a given task has been elusive, and most trained networks in use are accurate, black box predictors. We propose SAMPAT (Smooth Approximation using Multi-Polynomial and Analytic Transformations), a 3 layer neural architecture, that facilitates the construction of interpretable models and their analysis. Figure \ref{fig:arch1} describes the basic SAMPAT  architecture. 
    
    Inputs to layer 1 neurons are denoted by $x_1$, $x_2$, ... $x_n$ or $x_i$, $i = 1,$ 2, ..., $n$. The weight connecting input $x_j$ to the $i$-th first layer neuron is $u_{ij}$; the notation assumes the destination $i$ is the first letter of the subscript and $j$ is the source. The net input to the $i$-th first layer neuron, $net^1_i$ is given by
    \begin{gather}
     net^1_i = \sum_{j = 1}^n~ u_{ij} x_j + b_i
    \end{gather}
    
    where $b_i$ is a bias term that is added to the weighted sum. Without loss of generality, we assume that inputs include a constant input with a value of 1; a weight of $b_i$ associated with such an input serves the same purpose. Hence, unless required otherwise, we will drop the bias term to keep the notation and descriptions facile. First layer neurons have a logarithm activation function. The output of the $i$-th first layer neuron is denoted by $y_i$, and is given by
    \begin{gather}
     y_i = log(net^1_i) = log(\sum_{j = 1}^n~ u_{ij} x_j)
    \end{gather}
    The base of the logarithm is normally $e \approx 2.71828$, except for digital circuit implementations, where base 2 is desirable. Note that $log_2(net^1_i) = \frac{log_e(net^1_i)}{log_e(2)}$, and $log_e(net^1_i) = \frac{log_2(net^1_i)}{log_2(e)}$, implying that changing the base is equivalent to a gain term. Layer2 neurons receive weighted sums of layer 1 outputs. The weight of the connection from neuron $i$ in layer 1 to neuron $k$ in the layer 2 is denoted by $v_{ki}$. The net input to neuron $k$ in layer 2 is denoted by $net^2_k$ and is given by
        \begin{gather}
     net^2_k = \sum_{i = 1}^A~ v_{ki} y_i = \sum_{i = 1}^A~ v_{ki} ~log \left(\sum_{j=1}^n  u_{ij} x_j\right) = \sum_{i = 1}^A~ log \left(\sum_{j=1}^n  u_{ij} x_j\right)^{v_{ki}} = log \left( \prod_{i=1}^A ~ \left(\sum_{j=1}^n  u_{ij} x_j\right)^{v_{ki}} \right)
    \end{gather}
    where bias terms have been subsumed. Second layer neurons use an exponential activation function. The output of the $k$-th layer 2 neuron is denoted by $z_k$, and is given by $base^{net^2_k}$, where the “base” is identical to that used for the logarithm in layer 1. The base is again, normally $e$. Hence,
    \begin{align}
     z_k ~=~ exp(net^2_k) = exp \left( log \left( \prod_{i=1}^A ~ \left(\sum_{j=1}^n  u_{ij} x_j\right)^{v_{ki}} \right) \right) = \prod_{i=1}^A ~ \left(\sum_{j=1}^n  u_{ij} x_j\right)^{v_{ki}}
    \end{align}
    
    The output of the $k$-th second layer neuron is a reducible polynomial, composed of a product of polynomials of first layer inputs. Reducible polynomials of a single variable form a dense subset of the set of all polynomials, and a dense subset of the set of all smooth functions. Hence, a two layer SAMPAT network is capable of universal approximation of any continuous function of one variable, since a polynomial can be found that approximates the said smooth function as accurately as desired. However, the set of reducible multivariate polynomials is a thin subset of the family of multivariate polynomials, while the set of irreducible multivariate polynomials is a dense subset of the set of polynomials and of the set of continuous multivariate functions. Since any irreducible polynomial may be expressed as a linear combination of reducible polynomials, any irreducible polynomial may be computed by the use of a third layer that produces linear combinations of the reducible polynomials computed by the second layer. The input to the $l$-th third layer neuron is a weighted sum of second layer neuron outputs, given by
    
    \begin{gather}
     net^3_l = \sum_{k=1}^B~ w_{lk} z_k = \sum_{k=1}^B~ w_{lk} \prod_{i=1}^A ~ \left(\sum_{j=1}^n  u_{ij} x_j\right)^{v_{ki}}, \; l = 1, 2, ..., C.
    \end{gather}
    where we have again assumed that any bias terms are represented by a constant third level input whose associated weight equals the bias term. The output of the $l$-th third layer neuron is given by
    \begin{gather}
     t_l = act(net^3_l) = act \left( \sum_{k=1}^B~ w_{lk} z_k \right) = act \left( \sum_{k=1}^B~ w_{lk} \prod_{i=1}^A ~ \left(\sum_{j=1}^n  u_{ij} x_j\right)^{v_{ki}} \right), \; l = 1, 2, ..., C.
    \end{gather}
    where $act()$ is the activation function of a third layer neuron. In the case of regression tasks, $act()$ is usually the identity or a linear function, i.e. $act(input) = input$, or $act(input) = gain \cdot input$. 
    
    Hence, each third layer SAMPAT neuron can represent any multivariate polynomial, reducible or irreducible. The extended Stone-Weierstrass Theorem states that any continuous multivariate function can be approximated arbitrarily closely by a multivariate polynomial of sufficiently high degree. Hence, a three layer SAMPAT network can approximate any function as accurately as desired, by choosing appropriate weights for the three layers. Since polynomials are analytic functions capable of being differentiated any number of times, this means that SAMPAT representations provide all derivatives and partial derivatives. All SAMPAT representations can be expressed in algebraic form. In short, a three layer SAMPAT is capable of universal approximation, can provide all derivatives and partial derivatives, and full interpretability at any neuron of the network. The $log()$ function does not admit a zero argument, while negative arguments do not pose a hurdle if they are represented as polar forms involving complex variables, i.e. $log(\|r\|e^{i \theta}) = log(\|r\|) + c.p.v. (i \theta)$, where $c.p.v.$ denotes the Cauchy principal value. In a Methods Section, we discuss these aspects in more detail, including how real valued approximants may be computed without using complex arithmetic.\\
       
    \begin{figure}[ht!]
    \centering
    \begin{subfigure}[ht]{0.3\textwidth}
%         \centering
%         \includegraphics[width=0.3\linewidth]{figuv1-noma.png}
%         \caption{$(x_1 + x_2)^2$}
        \caption{\small $sin(x)$\\ $\approx 1.8\cdot(1.35x + 0.29)^{0.31} - 1.28$\\
        $\approx {0.67 + 1.56 \left(1.77 x\right)^{0.39} -1.69\left(1.77 x\right)^{0.13}}$
        } \label{figuv1}
%         \; R2 =  0.990415
    \end{subfigure}%
    ~ 
    \begin{subfigure}[ht]{0.3\textwidth}
        \centering
        \includegraphics[width=0.3\linewidth]{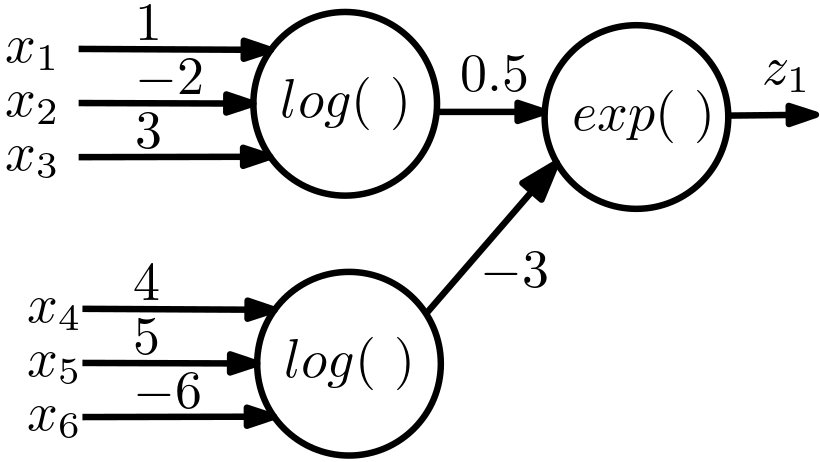}
         \caption{$\frac{(x_1 - 2x_2 + 3x_3)^{0.5}}{(4x_4 + 5x_5 - 	6x_6)^3}$}\label{figuv2}
    \end{subfigure}
     ~
     \begin{subfigure}[ht]{0.3\textwidth}
         \centering
         \includegraphics[width=0.3\linewidth]{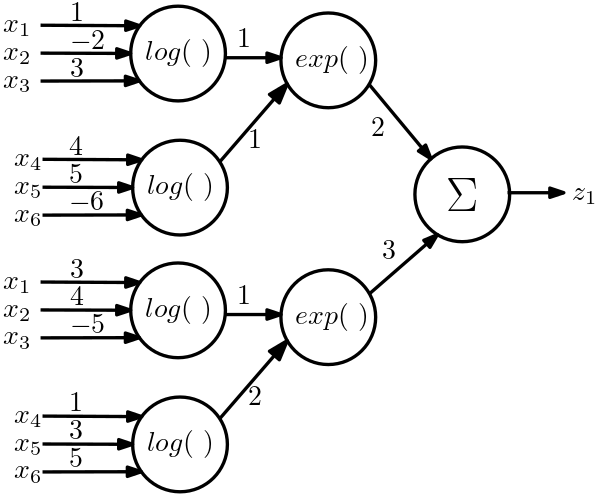}
          \caption{$2\frac{(x_1 - 2x_2 + 3x_3)}{(4x_4 + 5x_5 - 6x_6)} + 3 \frac{(3x_1 + 4x_2 - 5x_3)}{(x_4 + 3x_5 + 5x_6)^2}$}\label{figuvw7}
     \end{subfigure}
    \caption{(a) 1 and 2 term SAMPAT approximations to $sin(x)$; (b), (c) 2 and 3 layer SAMPAT networks for some functions.}
    \end{figure}
    
    Figures \ref{figuv2}, and \ref{figuvw7} show SAMPAT realizations for some illustrative examples. Alternative realizations are often possible. Connections between neurons have been restricted in these examples for illustration, but SAMPAT networks trained on data learn a suitable structure in which many issential weights are null or near zero, allowing for the discovery of a near optimal form of the approximant. Parsimonious approximations are valuable for hardware realization if they lead to small lookup tables.  Consider $sin(x)$ in the range $0$ to $\pi/2$. A one-term Taylor series approximation is $sin(x) \approx x$, which has a test R2 score of 0.76. Figure \ref{figuv1} shows SAMPAT approximations using one and two terms; both have R2 scores above 0.99.
%     
%     these were found by training 2 and 3 layer SAMPAT networks with 1 neuron in layer 1, 1 neuron in layer 2, and 1 third layer neuron, was trained by using 5000 training samples and 1000 test samples. It found the approximant
%     \begin{gather}
%     sin(x) \approx 1.8\cdot(1.35x + 0.29)^{0.31} - 1.28
%     \end{gather}
%     which has a test R2 score above 0.99.\\
%     $0.79x^{0.72}$, which yields a R2 score of 0.98. All models were trained with 3200 samples.

    The representational ability of a 3-layer SAMPAT can be further expanded by introducing skip connections, which connect an input, the net input to a neuron, or a neuron output,  to a neuron in a non-adjacent layer. When skip connections are present, the output of the $k$-th second layer neuron may be expressed as $z_k = exp(\alpha_k(x)) \prod_{i=1}^A ~ \left(\sum_{j=1}^n  u_{ij} x_j\right)^{v_{ki}}$, where $\alpha_k(x)$ is a function that depends on weights and inputs.
%     , and is detailed in supplementary files. 
    The use of skip connections increases the number of possible functions the network can represent or learn. 
    
%     When training neural networks, it is known that learning is simplified when multiple alternative representations are possible, since the training process is usually a gradient descent on a very nonlinear landscape with multiple minima. When multiple minima that are equally acceptable are present, the chances that a good representation is learnt by the network increase. Although SAMPAT networks tend to be very shallow, the use of skip connections also helps to back-propagate gradients faster to earlier layers, speeding up convergence.
    
    If the output of a 3 layer SAMPAT is provided as input to a neuron with an exponential activation function, the output of the latter can represent exponential functions of polynomials of input variables - which include Gaussians whose arguments are polynomials in input variables. A 6 layer SAMPAT with skip connections can represent polynomials, weighted sums of polynomial products, polynomials of the above, and mixtures of Gaussians whose exponents are weighted sums of polynomial products.
    
    The use of complex weights in SAMPAT significantly expands the family of functions that can be learnt or represented. Convergence during training is better, and less data suffices in many instances. Since real numbers are a subset of the set of complex numbers, all functions that can be computed by using real valued weights are a subset of the set of functions representable by complex SAMPAT networks. 
        
\begin{figure}[ht!]
    \centering
    \begin{subfigure}[ht!]{0.15\textwidth}
        \centering
        \includegraphics[width=\textwidth]{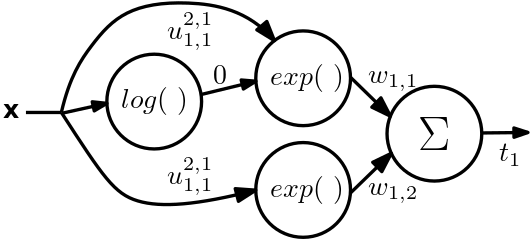}
        \caption{}\label{figuvwsine_trainable}
    \end{subfigure}%
    ~ 
    \hfill%
%     \caption{\tiny $\left(5.21328 \cdot 10^{-5} - 0.49987 i\right) e^{x \left(0.000153 + 0.99998 i\right)}$\\$ + \left(5.082099 \cdot 10^{-5} + 0.49987 i\right) e^{x \left(0.000153 - 0.99998 i\right)}$ \\ $ \approx \left( - 0.499i\right) e^{x \left(0.99998 i\right)} + \left(0.499 i\right) e^{x \left( - 0.99998 i\right)} \approx  \frac{e^{ix} - e^{-ix}}{2i}, i = \sqrt{-1}$}   
    \begin{subfigure}[ht!]{0.25\textwidth}
        \centering
        \includegraphics[width=\textwidth]{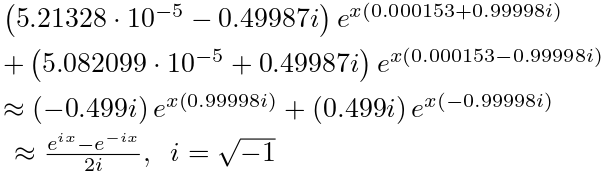}
        \caption{}
    \end{subfigure}%
    ~
    \hfill%
    \begin{subfigure}[ht!]{0.2\textwidth}
     \centering
    \includegraphics[width=\textwidth]{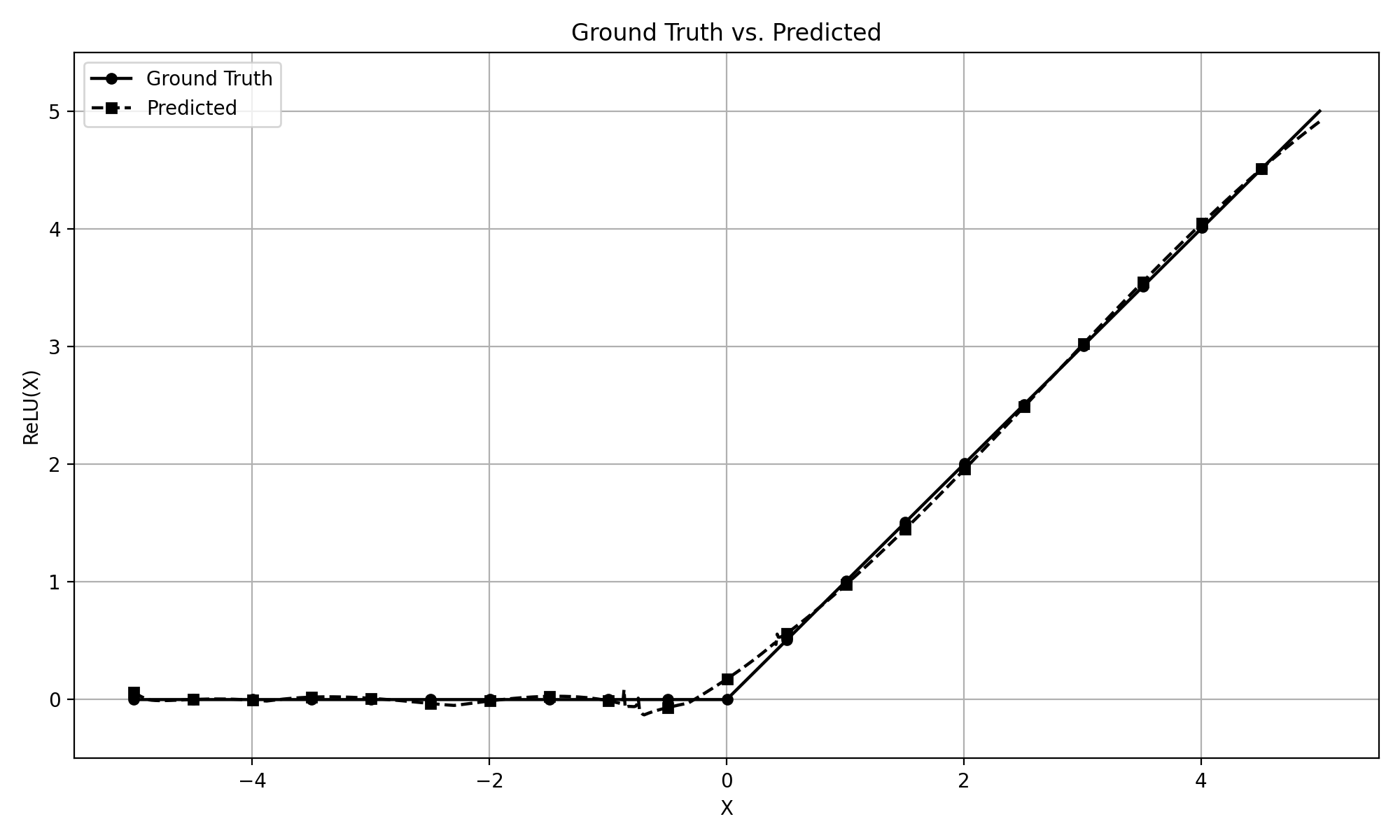}
    \caption{}\label{fig:sum_relu_eq}
%     (c)
    \end{subfigure}
    ~
    \hfill%
    \begin{subfigure}[ht!]{0.35\textwidth}
     \centering
    \includegraphics[width=\textwidth]{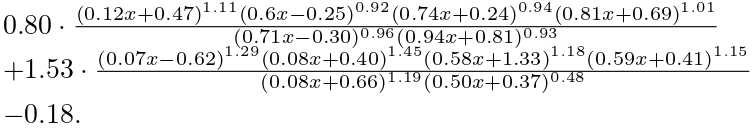}
% (d)
    \caption{}\label{fig:eqnrelu}
    \end{subfigure}    
    \caption{(a), (b): SAMPAT network and approximant for $sin(x)$; (c), (d): Approximant and plot for $Relu(x)$}\label{sinrelu}
\end{figure}

    When complex-valued weights are used, other representations emerge. Consider the SAMPAT network in in Fig.\ref{figuvwsine_trainable}. Weights whose values are indicated are indicated in the figure are fixed, while other weights are variable. Note that in this SAMPAT network, layer 1 weights are absent (fixed at 0), except for skip connections. The latter include $u^{2,1}_{1,1}$, which is a skip connection from input 1 of layer 1 to neuron 1 of layer 2, and $u^{2,1}_{2,1}$, which is the weight from input 1 of layer 1 to neuron 2 of layer 2. The only other weights that are learnable are weights $w_{1,1}$ and $w_{1,2}$ from the layer 2 neuron outputs to the single layer 3 neuron. The network was trained by using samples from $sin(x)$ function in the range $0 \leq x \leq 2 \pi$.

    Figures \ref{fig:sum_relu_eq} and \ref{fig:eqnrelu} shows the plot and approximant for $Relu(x) = max(x, 0)$, learnt by a 2 layer SAMPAT network with 1 and 6 neurons in the 2 layers, that was trained with 5000 training samples. This may be compared with approximants using 5, 9, and 13 degree Newman polynomials in \cite{telgarsky2017neural}.
    
%   \subsection{Factoring Polynomials}
    
\begin{figure}[ht!]
    \centering
    \begin{subfigure}[ht!]{0.25\textwidth}
        \centering
        \includegraphics[width=\textwidth]{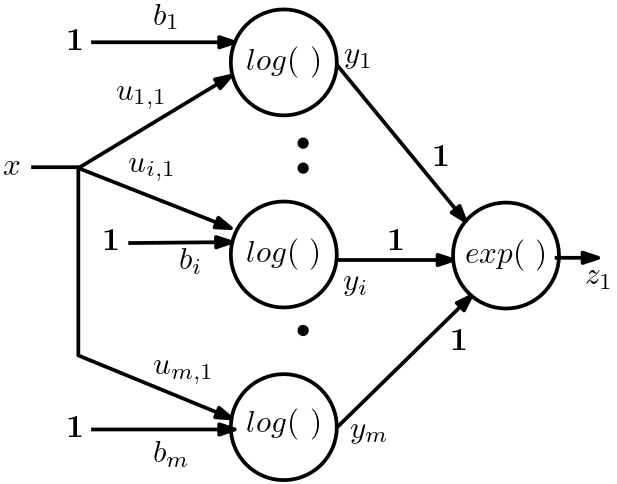}
%     \caption{Two layer SAMPAT with fixed layer 2 weights.}
    \caption{} \label{figunivariate_roots}
    \end{subfigure}%
    ~ 
    \hfill%
    \begin{subfigure}[ht!]{0.25\textwidth}
        \centering
        \includegraphics[width=\textwidth]{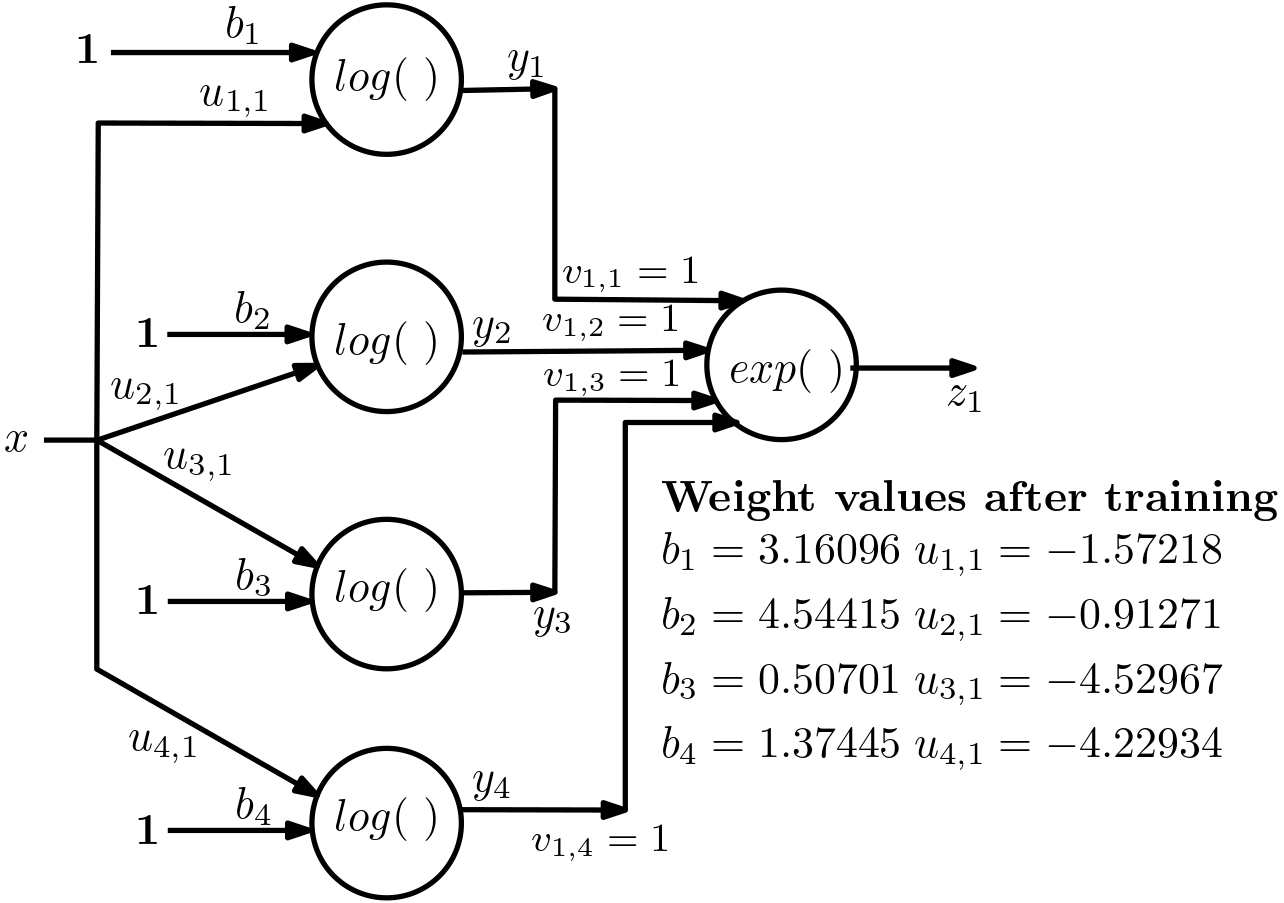}
%          \caption{SAMPAT network for determining roots of $x^4-19x^3+121x^2-309x+270$.}
\caption{}\label{figuv6}
    \end{subfigure}%
    \hfill%
    \begin{subfigure}[ht!]{0.2\textwidth}
        \centering
        \includegraphics[width=\textwidth]{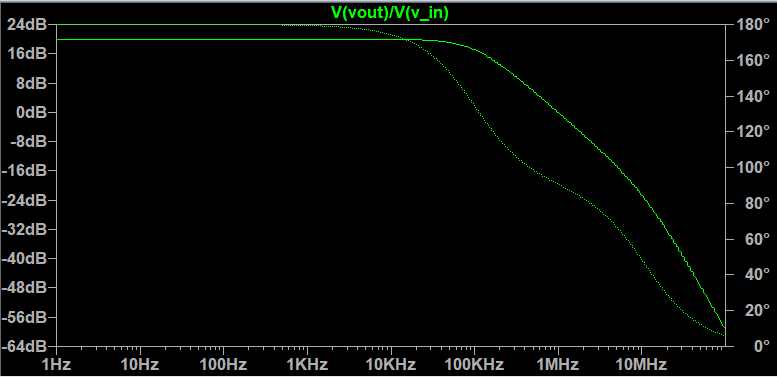}
    \caption{}\label{figuv7sim}
    \end{subfigure}%
    ~ 
    \hfill%
    \begin{subfigure}[ht!]{0.25\textwidth}
        \centering
        \includegraphics[width=\textwidth]{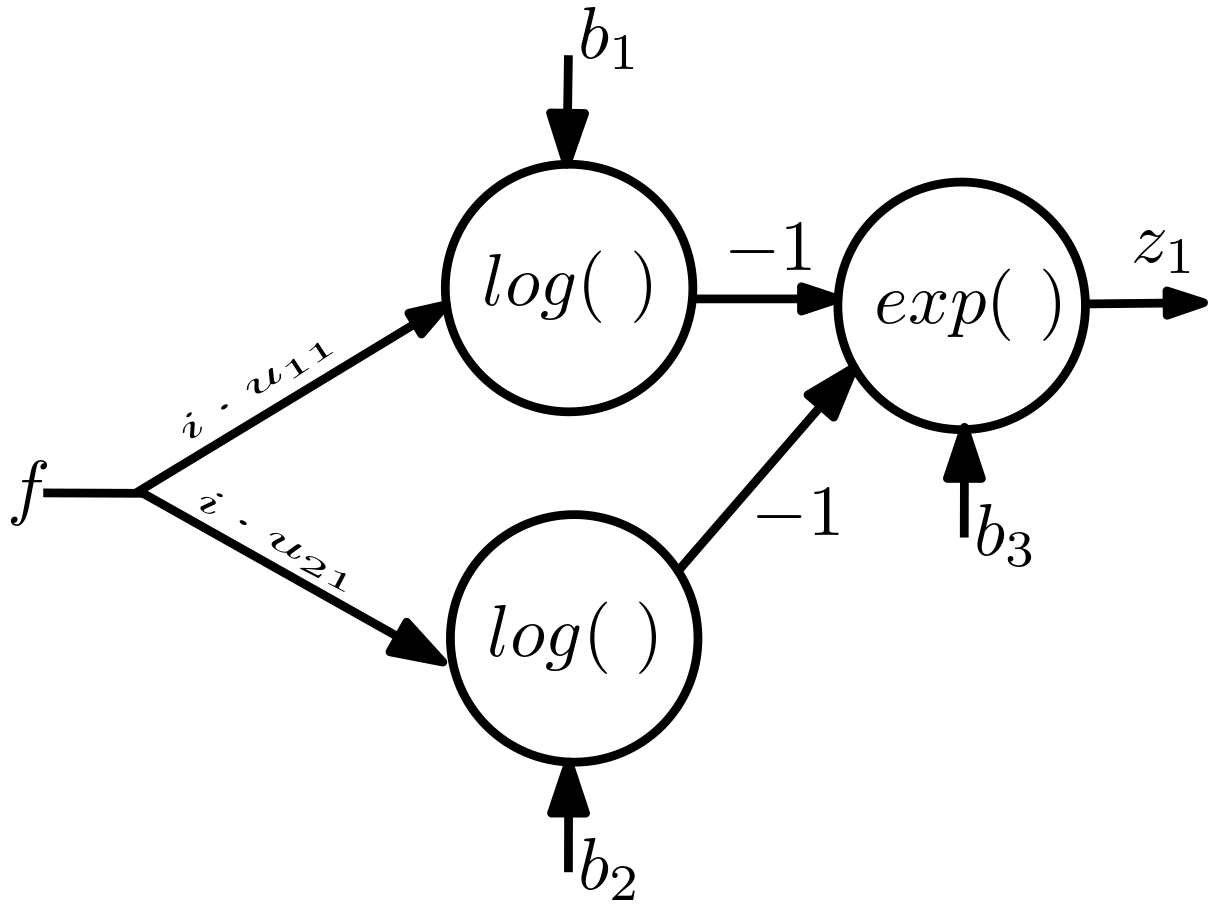}
         \caption{}\label{fig_bodeckt}
    \end{subfigure}%
    \hfill
    \caption{(a) SAMPAT for polynomial factoring and b) an example; (c) Op-amp frequency response, (d) SAMPAT model}
\end{figure}

    Consider the two layer SAMPAT shown in Fig. \ref{figunivariate_roots}, in which all second layer weights are equal to 1, and in which only one second layer neuron is present. The output of the layer 2 neuron, denoted by $z_1$, is given by
    \begin{align}
     z_1 = \prod_{i=1}^A ~ \left(\sum_{j=1}^n  u_{ij} x_j + b_i\right)
    \end{align}
    where we have re-introduced the bias term $b_i$ to facilitate the description. The bias term has been shown only for the first and for the $i$-th first level processing unit, for the sake of clarity.  

    Suppose that the weights $u_{ij}$ and bias terms $b_i$ are complex valued, but all layer 2 weights  $v_{1i}$ are equal to 1. In this case, $\left(\sum_{j=1}^n  u_{ij} x_j + b_i\right)$ are complex roots of $z_1$. Fig. \ref{figuv6} shows a SAMPAT network for factoring $f(x) = x^4-19x^3+121x^2-309x+270$, along with weight values after training using 1000 samples. The output
    \begin{align}
     z_1 = 1.0 \left(3.16096 - 1.57218 x\right)^{1.0} \left(4.54415 - 0.912706 x\right)^{1.0} \left(0.507008 x - 4.52967\right)^{1.0} \left(1.37445 x - 4.22934\right)^{1.0}
    \end{align}
    from which the roots may be read as $x = 8.934,3.077,4.978,2.01$, that are close to the exact roots \textit{viz.} $x = 9,3,5,2$. Consider $f(x) = x^3 - 1$, whose factors are the cube roots of unity. We use a 2 layer SAMPAT with complex weights, 1 and 3 neurons in layers 1 and 2, with all input weights of the layer 2 neuron fixed at 1. The approximant learnt after 15,000 training epochs on 1000 samples yields
    \begin{gather}
     z_1 = 1.0 \left(1.0 x - 0.9999 + 1.1239 \cdot 10^{-6} i\right)^{1.0} \left(1.0 x + 0.4999 - 0.8660 i\right)^{1.0} \left(1.0 x + 0.4999 + 0.8660 i\right)^{1.0}
    \end{gather}
    On ignoring negligible weights, the roots are then given by $x=1,-0.4999 - 0.866 \cdot i,-0.4999 + 0.866 \cdot i$, that closely approximate the exact roots given by $x=1,-\frac{1}{2}+\frac{\sqrt{3}}{2}\cdot i,-\frac{1}{2}+\frac{\sqrt{3}}{2}\cdot i$.\\
    
%     \subsection{System Modelling}
    
    In electronics, signal processing, and control engineering applications, there is a frequent need to model the frequency response of a nonlinear system. We consider an operational amplifier, which is often modelled by using a single or two-pole model. The operational amplifier circuit was simulated by using the LTSpice circuit simulator \cite{LTspice2026}, to obtain the magnitude and phase responses; an illustrative example is shown in Fig. \ref{figuv7sim}.

%     \begin{figure}[ht!]
%     \centering
%     \begin{subfigure}[ht!]{0.6\textwidth}
%         \centering
%         \includegraphics[width=\textwidth]{opamp_cktsim.png}
%     \caption{Magnitude and phase responses of an operational amplifier circuit obtained by using a circuit simulator.}\label{figuv7sim}
%     \end{subfigure}%
%     ~ 
%     \hfill%
%     \begin{subfigure}[ht!]{0.35\textwidth}
%         \centering
%         \includegraphics[width=\textwidth]{fig_bodeckt.png}
%          \caption{SAMPAT network for a 2 pole op-amp model.}\label{fig_bodeckt}
%     \end{subfigure}%
%     \hfill
%     \caption{Two layer SAMPAT for learning operational amplifier models.}
% \end{figure}
% 
%     \begin{figure}[ht]
%     \centering
%     \includegraphics[width=0.5\linewidth]{opamp_cktsim.png}
%     \caption{Magnitude and phase responses of an operational amplifier circuit obtained by using a circuit simulator.}
%     \label{figuv7sim}
%     \end{figure}
    
    The frequency response (or input-ouput transfer function) of the operational amplifier is usually modelled by using an all-pole models with one or two poles. A two pole model assumes the form
    \begin{equation}
    \label{eq:opamp_tf}
    H(s) = \frac{A_0}{(1 + s/\omega_1)(1 + s/\omega_2)},
    \end{equation}

    where $A_0$ is the DC gain and $\omega_1$, $\omega_2$ are the two pole frequencies. Identifying $\omega_1$ and $\omega_2$ from measured frequency response data (the Bode plot) is a core task in analog circuit design and verification. In this case, note that the complex variable $s = i\omega = i 2\pi f$, which lies on the imaginary axis. We use the two layer SAMPAT network shown in Fig. \ref{fig_bodeckt}. The two first neurons have weights $i \cdot u_{11}$ and $i \cdot u_{21}$, which are both imaginary; their biases $b_1$ and $b_2$ have been fixed at 1, and the bias of the single layer 2 neuron, denoted by $b_3$ is learnable. The output $z_1$ may be written as
    \begin{gather}
      z_1 = \frac{exp(b_3)}{\left(1 + i\cdot u_{11}\right) \cdot \left(1 + i\cdot u_{21}\right)}
    \end{gather}
    
    Data was obtained by simulating the LTspice UniversalOpamp2 macromodel. The model is parametric, and we prescribed poles at 10 MHz and 20 MHz. This violates the commonly used dominant pole approximation; the poles intereact with each other, making modelling difficult. A set of 4001 logarithmically spaced samples were chosen in the frequency range of 100kHz to 1 GHz. The SAMPAT network was trained with the sum of the MSE lossses on the magnitude and phase response. The weights of the neural network after training yield the approximant
    \begin{gather}
    H(f) =
\frac{-9.98681 + 0.682064i}{\left[1 + ix(0.9644796)\right] \left[1 + ix(0.36026403) \right]} \text{, where } x = 10^{-7}~f.
%     H(s) \approx \frac{10.0} {\left(\frac{e^{-1.498\pi i}}{11.098{\times}10^6}x + 1\right) \left(\frac{e^{0.51\pi i}}{97.81{\times}10^3}x + 1\right)}
    \end{gather}
    
    A two pole approximation using the prescribed pole locations of 10 MHz and 20 MHz has a R2 score of 0.98. SAMPAT learns a model with pole locations at 10.368 MHz and 27.757 MHz, achieving a test $R2 \approx 0.9999$. SAMPAT's approximation is better because the circuit simulator uses a nonlinear op-amp model, and the pole prescriptions very loosely approximate the response.
    
%     Figures \ref{figuv7mag} and \ref{figuv7phase} show plots of the true and predicted magnitude and phase responses based on the model. A conventional modelling approach would use a dominant pole approximation, in which the second pole is assumed to lie at least a decade away from the first one. SAMPAT obviates the need for such assumptions.
%     \begin{figure}[ht]
%     \centering
%     \includegraphics[width=0.5\linewidth]{opamp_mag.png}
%     \caption{True and predicted magnitude responses of an operational amplifier model.}
%     \label{figuv7mag}
%     \end{figure}
%     
%     \begin{figure}[ht]
%     \centering
%     \includegraphics[width=0.5\linewidth]{opamp_phase.png}
%     \caption{True and predicted phase responses of an operational amplifier model.}
%     \label{figuv7phase}
%     \end{figure}
    
    Thus far, we have expanded on the use of SAMPAT for univariate functions. Functions of many variables, real or complex, can also be efficiently learnt or represented by SAMPAT. Table \ref{tab:multivariate} shows that SAMPAT can learn better approximations on multivariate functions, in comparison to an optimized conventional neural network (denoted as NN$_{\rm Optim}$), while using 7X-8X fewer parameters.
    
  \begin{table}[H]
  \centering
  \caption{Multivariate function approximation with SAMPAT}
  \label{tab:multivariate}
  \resizebox{\linewidth}{!}{%
  \begin{tabular}{lcccc}
    \toprule
    \textbf{Target Function} & \textbf{NN$_{\rm Optim}$ MSE}
      & \textbf{SAMPAT MSE} & \textbf{NN$_{\rm Optim}$ Params} & \textbf{SAMPAT Params} \\
    \midrule
    $y = -8x_1 + 5{\times}10^{-6}(1-e^{2x_2+2x_3}) + 2x_4^2$
      & 1.1337 & \textbf{0.7654} & 1110 & \textbf{161} \\
    $y = -8x_1^2 + e^{-6}\frac{2x_2}{2x_3} + e^{-6}(2x_4)(7x_5)$
      & 1.2374 & \textbf{0.6162} & 1245 & \textbf{171} \\
    \bottomrule
  \end{tabular}}
\end{table}
    
%     \subsection{System Identification}
    A number of properties of SAMPAT combine to make it especially useful for identifying systems from time series measurements. Consider a $2^{nd}$ order series RLC circuit. Depending on the values of R, L, and C, the roots may be both real, or both complex. The governing equation of the circuit is
    \begin{gather}
     \ddot{i}(t) = A \dot{i} + B i(t) + C.
    \end{gather}
    where $i(t)$ is the branch current. We use a 3 layer SAMPAT to fit samples of $i(t)$ as a function of time $t$. From the interpretable form of $i(t)$ thus obtained, we analytically determine $\dot{i}(t)$ and $\ddot{i}(t)$, which are then used to determine a linear relation between the three. In more complex examples, a second SAMPAT network may be used to learn an interpretable relation that represents the governing equation of a system. Table \ref{TableRLC} shows the identified system for an underdamped and an overdamped circuit.
    
  \begin{table}[H]
  \centering
  \caption{Summary of recovered governing equations for the three RLC cases.}
  \label{TableRLC}
  \resizebox{\linewidth}{!}{%
  \begin{tabular}{llllrrr}
    \toprule
    \textbf{Case} & \textbf{True ODE} & \textbf{True normalized form}
    & \textbf{Learned normalized form}
    & \textbf{$R^2(i)$} & \textbf{$R^2({\dot{i}})$} & \textbf{$R^2({\ddot{i})}$} \\
    \midrule
    Overdamped &
    $2\ddot{i}+4\dot{i}+2i=0$ &
    $\ddot{i}=-2\dot{i}-i$ &
    $\ddot{i}=-1.88298\dot{i}-0.92087i+0.000731717$ &
    $0.99999440$ & $0.99996014$ & $0.99297175$ \\
    Underdamped &
    $10\ddot{i}+2\dot{i}+10i=0$ &
    $\ddot{i}=-0.2\dot{i}-i$ &
    $\ddot{i}=-0.199894\dot{i}-1.00009i-0.000103023$ &
    $0.99999973$ & $0.99999988$ & $0.99999983$ \\
%     General &
%     $\ddot{i}+2\dot{i}+3i=0$ &
%     $\ddot{i}=-2\dot{i}-3i$ &
%     $\ddot{i}=-2.0005\dot{i}-3.00085i+0.000668382$ &
%     $0.99997923$ & $0.99999995$ & $0.99999988$ \\
    \bottomrule
  \end{tabular}}
\end{table}
    
%     shown in Fig. \ref{figrlc}. 
%     \begin{figure}[ht!]
%      \centering
%      \includegraphics[width=0.3\linewidth]{rlc1.png}
%     \caption{Series RLC circuit.}\label{figlc}
%     \end{figure}
%     A more complex example of a system of coupled differential equations is detailed in supplementary files. The dynamics of the Lorenz system and its learnt approximant are illustrated in Fig. \ref{fig_rossler}.
%     
    Other interesting examples are included in the supplementary files. Suitable restrictions on connections constrains SAMPAT networks to learn models that are rational functions, sums of rational functions, etc. However, in most cases, if no restrictions are imposed, the network learns an optimal or near optimal form, that largely depends on the data provided. When the inputs and/or the weights are complex, the input-output map may be used to achieve a filtering operation. It is possible to design nonlinear filters that use very few parameters and yet outperfporm their linear counterparts. Some examples are provided in the sequel. 
%     In supplementary files, we provide a gallery of examples that illustrate how SAMPAT can be used to learn a diversity of approximants for a multitude of functions.\\
    
% \subsection{Classificaton, Regression, Feature Selection}
The ability to learn from a very large family of smoooth functions makes SAMPAT amenable to regression tasks. This also enables an arbitrarily complex decision boundary to be learnt. Choosing sigmoid or $tanh()$ functions for the $act()$ activation function in the 3rd layer allows SAMPAT to be directly used for classification tasks. 

\begin{figure}[ht!]
    \centering
    \begin{subfigure}[ht!]{0.35\textwidth}
        \centering
        \includegraphics[width=\textwidth]{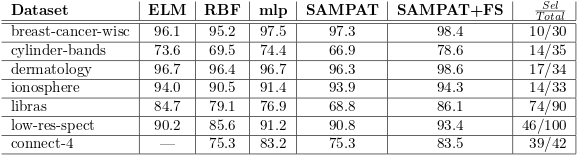}
    \caption{} \label{classtable}
  \end{subfigure}%
   \hfill%
    \begin{subfigure}[ht!]{0.25\textwidth}
        \centering
        \includegraphics[width=\textwidth]{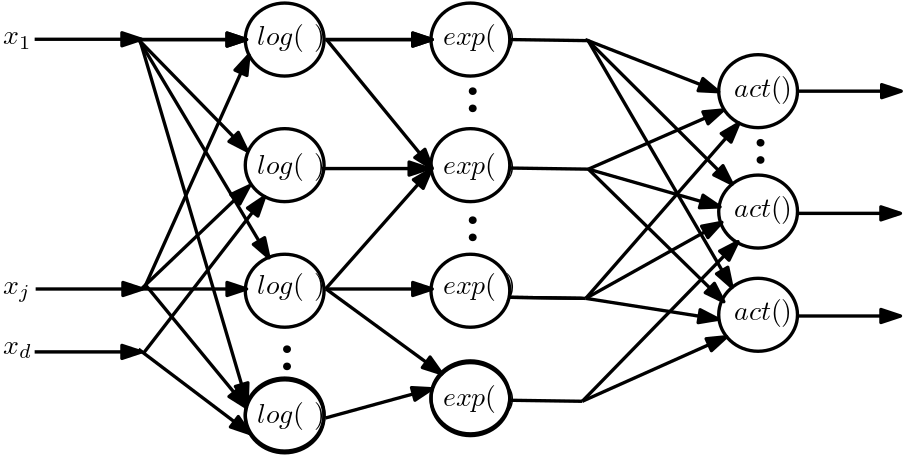}
        \caption{} \label{sampat-fs}
  \end{subfigure}%
  \hfill%
  \begin{subfigure}[ht!]{0.30\textwidth}
        \centering
        \includegraphics[width=\textwidth]{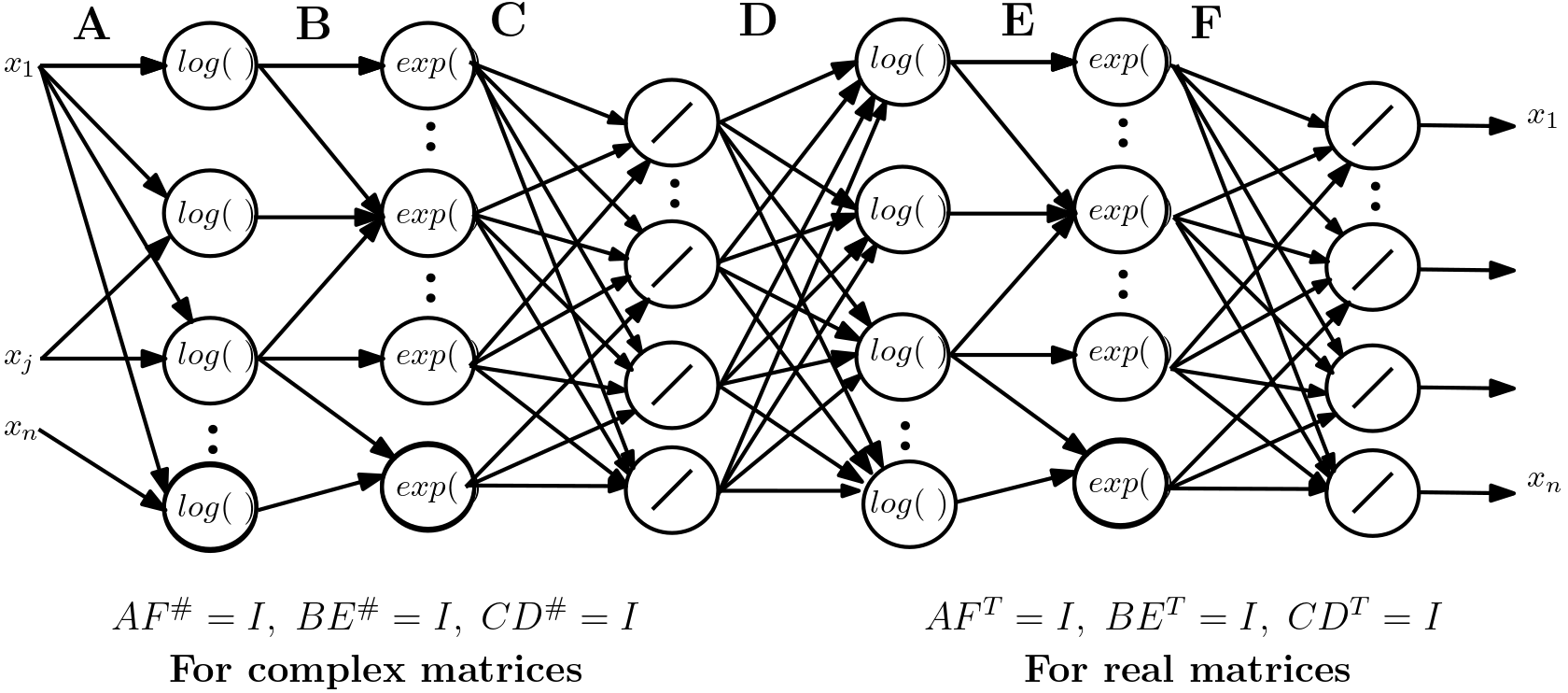}
        \caption{} \label{sampat2block}
  \end{subfigure}%
  \caption{(a) SAMPAT network for feature selection (b) Classification benchmark results (c) 2 SAMPAT blocks } 
\end{figure}

Figure \ref{classtable} presents illustrative results on a few benchmark datasets from the UCI repository, including comparisons with some well known methods, \textit{viz. } the Extreme Learning Machine (ELM) \cite{huang2006extreme}, Radial basis function SVMs, and an optimized multilayer feedforward neural network. Figure \ref{sampat-fs} shows a SAMPAT network with an additional diagonal weight layer before layer 1. This additional layer does not augment the network's capacity. Regularizing this layer's weights accomplishes feature selection; features are combined in nonlinear ways in later layers to determine the regressor or classifier. The column SAMPAT+FS in Fig. \ref{classtable} shows results with such a feature selection step, while the last column indicates the number of selected features vs. the total number of features. Little or no tuning was done to optimize results with SAMPAT, and these results are provided for illustration only.\\

\section{Disscussion}
\begin{figure}[ht!]
    \centering
    \begin{subfigure}[ht!]{0.22\textwidth}
        \centering
        \includegraphics[width=\textwidth]{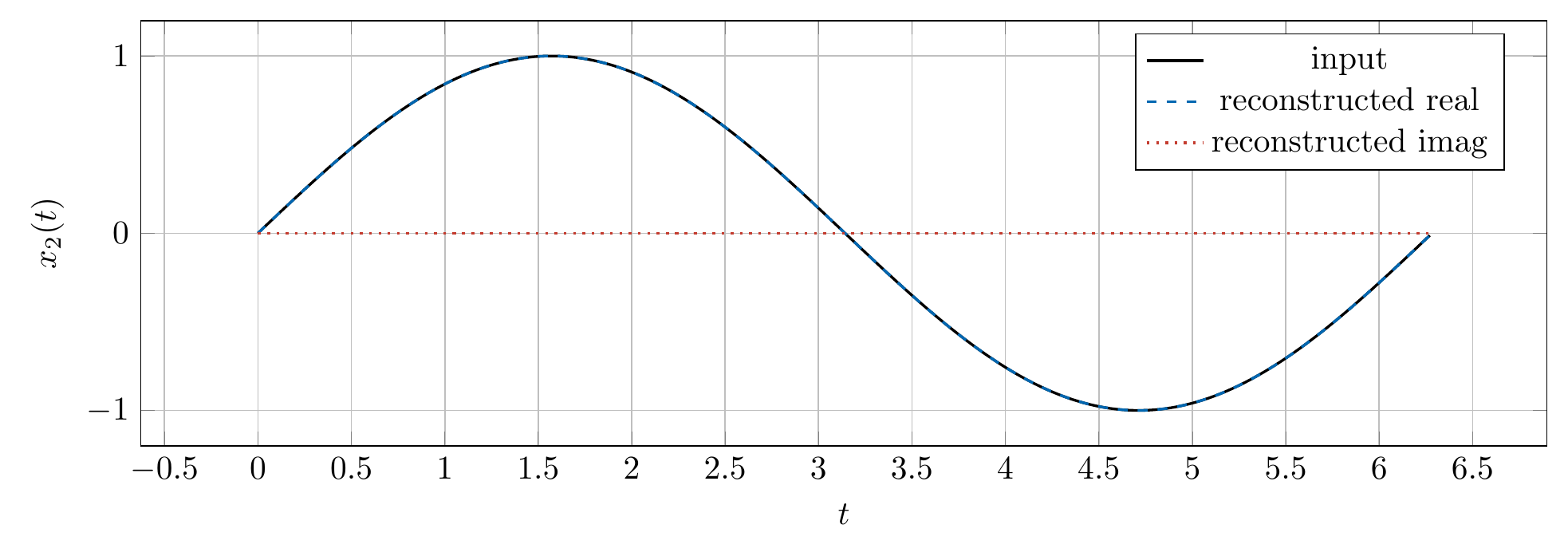}
    \caption{}\label{tx1sine}
  \end{subfigure}%
   \hfill%
    \begin{subfigure}[ht!]{0.22\textwidth}
        \centering
        \includegraphics[width=\textwidth]{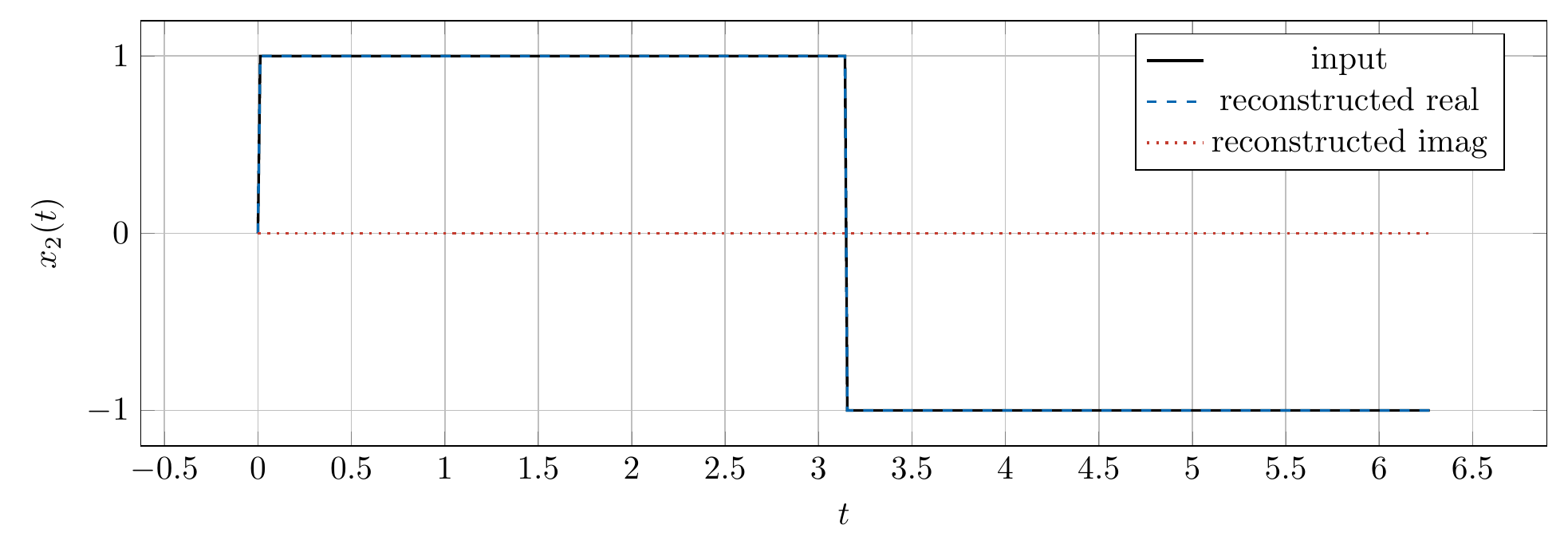}
        \caption{}\label{tx1sq}
  \end{subfigure}%
    \begin{subfigure}[ht!]{0.22\textwidth}
        \centering
        \includegraphics[width=\textwidth]{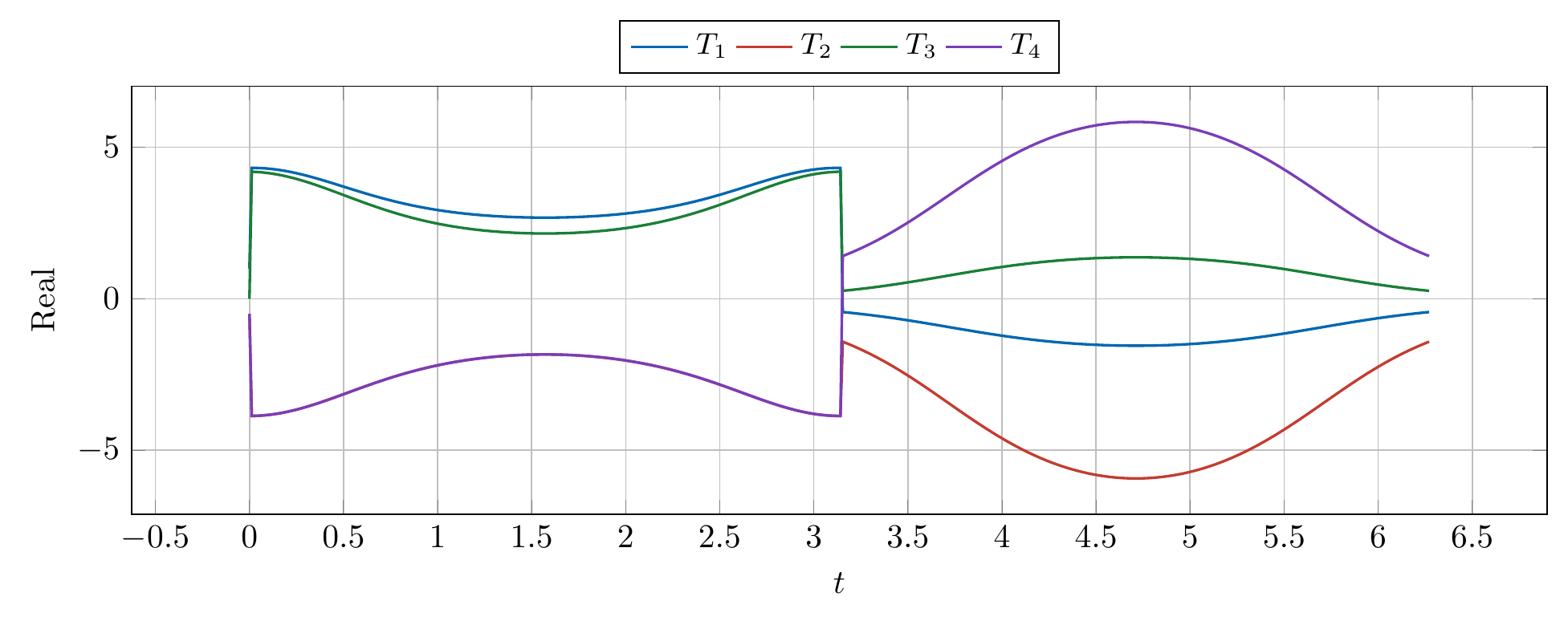}
    \caption{}\label{uvw1real}
  \end{subfigure}%
   \hfill%
    \begin{subfigure}[ht!]{0.22\textwidth}
        \centering
        \includegraphics[width=\textwidth]{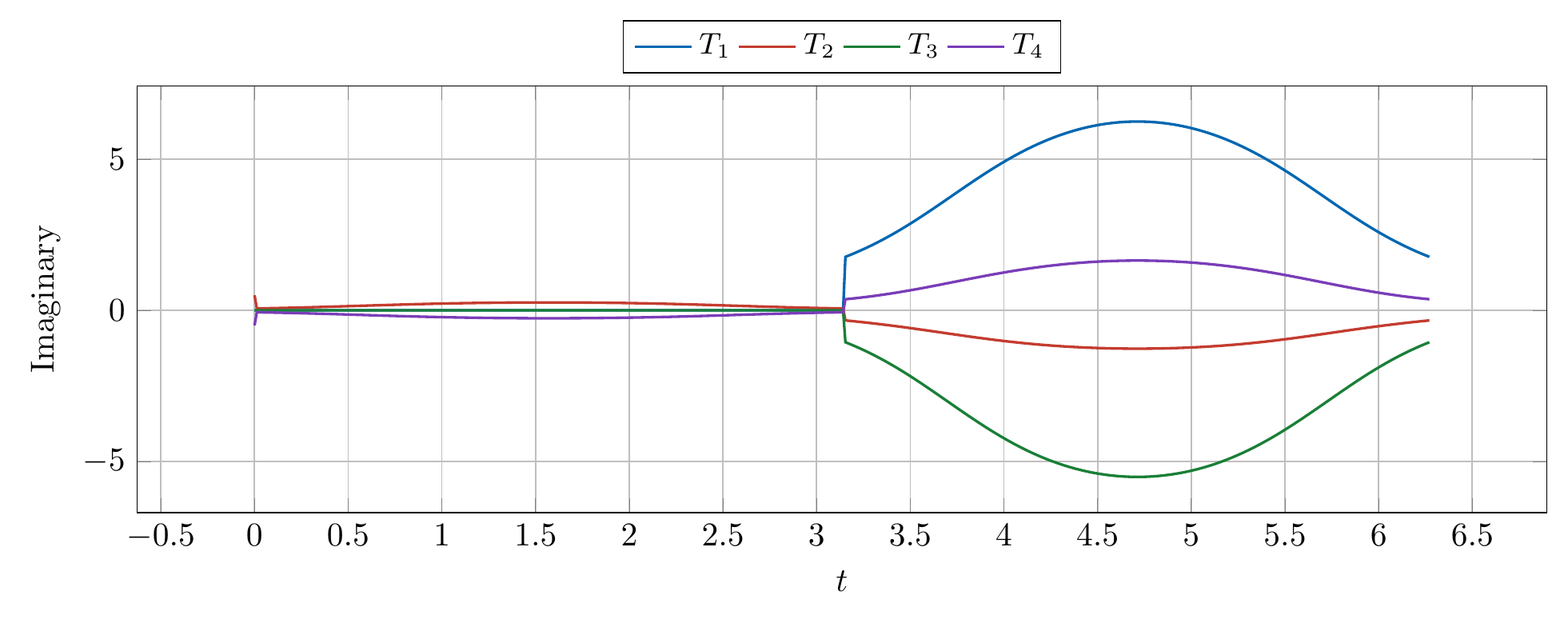}
        \caption{}\label{uvw1imag}
  \end{subfigure}%  
  \caption{(a), (b): Square and sine wave inputs to a $1^st$ SAMPAT block; (c), (d): Real and imaginary outputs of the first SAMPAT network; (a), (b) also show reconstructed outputs from the second SAMPAT network superimposed on the inputs.} 
\end{figure}

Specifying a family of approximants enables parameters to be optimized for a given application; well known examples includes Butterworth, Chebyshev, and rational functions \cite{chok2025rational, morina2025mathcal}. SAMPAT allows the family to be determined as part of the optimization process. The three layer structure of SAMPAT facilitates analysis and synthesis. Consider two SAMPAT networks, each of 3 layers. We denote the weight matrices of the first as A, B, and C, and of the latter as D, E, and F, respectively, as shown in Fig. \ref{sampat2block}. If the matrix pairs $(A, F)$, $(B, E)$ and $(C, D)$ are conjugate transposes of each other, i.e. 
\begin{gather}
 CD^\# = I \text{, } BE^\# = I \text{, } AF^\# = I \label{eqninv},
\end{gather}
then the output of the second SAMPAT block is an exact reconstruction of the input signals, as can be seen from Figs. \ref{tx1sine} and \ref{tx1sq}. Here, $I$ is an identity matrix of appropriate dimension. $Q^\#$ denotes the conjugate transpose of a matrix Q. In the case of real matrices, the relation (\ref{eqninv}) holds true when the conjugate transpose is replaced by the transpose, i.e. $CD^T = I$, $BE^T = I$, and $AF^T = I$. A nonlinear transformation and its approximate inverse are also possible, when
\begin{gather}
 CD^\# \approx I \text{, } BE^\# \approx I \text{, } AF^\# \approx I \label{eqninvapprox}
\end{gather}
Relations for networks with skip connections are easily derived. Note that the discrete Fourier and wavelet transforms relate to networks with only skip connections. Consider an example in which $A$, $B$, and $C$ are $2 \times 3$, $3 \times 4$, and $4 \times 4$ complex valued matrices, i.e. the first SAMPAT block has 4 output neurons. Figures \ref{tx1sine} and \ref{tx1sq} show signals fed to the first SAMPAT block. Real and Imaginary components of the 4 neuron outputs of the first network are shown in Figs. \ref{uvw1real} and \ref{uvw1imag}. These form inputs to the second SAMPAT block. Outputs of the second SAMPAT block are shown in Figs. \ref{tx1sine} and \ref{tx1sq}, superimposed on the input signals.

% In supplementary files, we show an example in which a sinewave and a square wave are mixed nonlinearly, and then perfectly recovered.\\

Note that a large class of smooth functions may be represented by the weights of a SAMPAT network. The facility accorded by a parametric and interpretable of functions facilitates the understanding of many phenomena in deep learning, such as double descent. Table \ref{Tabledeep1} summarizes results obtained by using SAMPAT+ networks, which combine convolutional layers that use $Relu()$ activations, with convolutional layers that use the activations used in SAMPAT networks. Table \ref{Tabledeep1} shows results on the CIFAR10 \cite{parkhi2012cats} and CatsVsDogs \cite{parkhi2012cats} datasets. , which we denote as a SAMPAT+ network. Comparisons are provided with traditional networks achieving similar test set accuracies \cite{huang2017densely, team2023keras}. SAMPAT based networks were trained \textit{ab initio}, unlike conventional deep architectures that were typically pretrained on larger datasets like ImageNet, and then fine tuned on CIFAR10 and CatsvsDogs datasets. While these results use innovations motivated by SAMPAT networks. These efforts are at a preliminary level, and work to extend these ideas to conventional deep architectures is in progress.

It has not escaped our attention that SAMPAT may find use in the design and applications of new transformations and their inverses, novel ways of compression and dimensionality reduction, encryption, the extraction of components from mixtures, as well as in differentiable simulation and quantum signal processing. In primate vision, processing starts with photoreceptors computing the logarithm of incident light intensity, followed by linear LGN layers, and V1 neurons with power law activation functions. Some of our ongoing modelling work indicates that SAMPAT can explain phenomena like divisive normalization \cite{carandini1997linearity}, that has emerged as a canonical neural computation.

% In supplementary files, we provide our trained models for the CIFAR10 and CatsvsDogs datasets, for the interested reader.\\

\begin{table}[H]
  \centering
  \caption{Deep Convolutional SAMPAT+ networks: early results}\label{Tabledeep1}
  \begin{tabular}{lcc}
    \toprule
    \textbf{Dataset} & \textbf{Test Accuracy} & \textbf{Inference Parameters}\\
    \midrule
    CIFAR10 & {90.66\% (SAMPAT+) ~~ 90.49\% (DenseNet)} & {89,366 (SAMPAT+) ~ 800,000 (DenseNet)}\\
    CatsvsDogs & {97.91\% (SAMPAT+) ~~ 98.38\% (MobileNet V2)} & {124,482 (SAMPAT+) ~~ 3.5M (MobileNet V2)}\\
    \bottomrule
  \end{tabular}
\end{table}

\section{Acknowledgement}
The authors are would like to thank Mr. Anil Pahal, Professors Amit Kumar, Basabi Bhaumik, Suresh Chandra, Pulin Kumar Bhattacharyya, and Mr. Arjun Sai Krishnan for valuable comments and useful discussions. This work was supported by project FT/2024/03/05 at FITT, IIT Delhi, funded by Sparsemind Technology Labs Pvt. Ltd.
  
\bibliographystyle{IEEEtran}
\bibliography{sampat_arxiv.bib}
\newpage
\vfill
\appendix
\section{Appendix I}

%     \begin{figure}[ht]
%     \centering
%     \includegraphics[width=0.5\linewidth]{uvwsine.png}
%     \caption{A manual construction illustrating a realization of the $sin()$ function.}
%     \label{uvwsine}
%     \end{figure}
%     In this case, the output is given by
%     \begin{align}
%      exp(\sqrt{-1} \cdot x) \cdot (-0.5 \sqrt{-1})  - exp(-\sqrt{-1} \cdot x) \cdot (0.5 \sqrt{-1}) 
%      = \frac{e^{\sqrt{-1} \cdot x} - e^{-\sqrt{-1} \cdot x}}{2 \sqrt{-1}}
%     \end{align}
%  In the range $0$ to $\pi$, a two term Taylor series expansion is $\left(x - \frac{x^3}{3!}\right)$, which has a R2 score of 0.24. A 3 layer SAMPAT network was used, with with 1, 2, and 1 neurons in layers 1, 2, and 3, respectively. All models were trained with 3200 training samples and tested on 800. The 3 layer SAMPAT network yields the approximant $\left[1.7 x^{1.2} - 0.84 x^{1.8}\right]$, which has a R2 score of 0.99.

\subsection{Approximations to $sin(x)$}
    For the range $0$ to $2\pi$, we used 6400 training samples to train a 3 layer SAMPAT network, with 1, 5, and 1 neurons in layers 1, 2, and 3 respectively. The 5 term Taylor series expansion is $\left(x - \frac{x^3}{3!} + \frac{x^5}{5!}- \frac{x^7}{7!} + \frac{x^9}{9!}\right)$, which has a R2 score of -11.914186. The SAMPAT network learns the approximant
    \begin{gather}
     -0.380352 x^{0.455587} + 2.18423 x^{1.09932} - 0.601864 x^{2.06027} - 0.502569 x^{2.10766} + 0.191451 x^{2.83966} \nonumber\\
    \end{gather}
    which has a test R2 score of 0.993. 
    
%     Figure \ref{uvw2} shows a plot of the ground truth, together with the Taylor and 3 layer SAMPAT regressors.
%     \begin{figure}[ht]
%     \centering
%     \includegraphics[width=0.5\linewidth]{uvw2.png}
%     \caption{The function $sin(x$ for $0 \leq x \leq 2 \pi$, along with a 3 layer SAMPAT and 5 term Taylor series regressors.}
%     \label{uvw2}
%     \end{figure}
    
%     In these cases, the second layer weights $v_{ki}$ were constrained to be positive, so that all terms have positive powers. If this constraint is removed, other expansions can also be learnt. For example, in the input range $1 \leq x \leq 4$, a 3 layer SAMPAT network with 2, 2, and 1 neurons in layers 1, 2, and 3, respectively, finds the approximant
%     \begin{align}
%      \frac{4.67}{\left(1 - 0.15 x\right)^{0.62} \left(- 0.04 x - 1\right)^{1.47}} -\frac{1.54 \left(- 0.04 x - 1\right)^{1.0}}{\left(1 - 0.15 x\right)^{1.32}} - 1.8
%     \end{align}

\subsection{Factoring Polynomials}
Another example is the 7th degree poynomial function $f(x) = x^7-9.5x^6+29.5x^5-27.5x^4-18.5x^3+37x^2-12x$. This function may be factored as $f(x) = (x - 4) (x - 3) (x - 2) (x - 1) (x - 0.5) x (x + 1)$. We use a two layer SAMPAT network with 7 and 1 neurons in the first and second layers, respectively. First first layer inputs have complex valued weights; all second layer weights are fixed at 1.0. The network was trained over 30,000 epochs with 1000 training samples. The neural network learnt the approximant
    \[
    \begin{split}
    1.0 \left(1.166 x e^{- 2.6161i} + 0.582 e^{0.525i}\right)^{1.0} &\left(5.476 x e^{- 1.592i} + 5.480 e^{1.549i}\right)^{1.0}\\ \left(0.785 x e^{- 0.593i} + 1.5702 e^{2.547i}\right)^{1.0} &
    \left(0.264 x e^{- 0.504i} + 1.055 e^{2.636 i}\right)^{1.0} \\
    \left(1.24134 x e^{0.610i} + 8.1411 \cdot 10^{-5} e^{- 2.479i}\right)^{1.0} &
    \left(0.714 x e^{2.321 i} + 2.142 e^{- 0.820 i}\right)^{1.0} \left(0.852 x e^{2.375 i} + 0.852 e^{2.375 i}\right)^{1.0}
    \end{split}
    \]
    where $i$ denotes $\sqrt{-1}$. From the factors of the expression, the zeros of $f(x)$ are approximately given by
    \begin{align}
     x=-0.499e^{3.141i},-1.000e^{3.141i},-2e^{3.14i},-3.996e^{3.14i},0,-3e^{-3.141i},-1e^{0i}
    \end{align}
    where $i$ denotes $\sqrt{-1}$. Since $\pi \approx 3.14$, $cos(\pi) = -1$, $sin(\pi) = 0$, $cos(0) = 1$, $sin(0) = 0$, and $e^{i\theta} = cos(\theta) + i sin(\theta)$, the roots may be written more simply as
    \begin{align}
     x=0.499,1.000, 2, 3.996, 0, 3, -1
    \end{align}
    which are fairly close to the true roots.\\
    
%     \begin{gather}
%      (1 - \frac{f_{pred}}{f_{actual}})^2 + (1 - \frac{g_{pred}}{g_{actual}})^2 \\
%      \text{where } g = \frac{1}{f}
%     \end{gather}
% 
%     \begin{gather}
%     H(s) = \frac{1.0}{
%     \left(2.84723{\times}10^{-8} e^{-1.73762i}s + 0.316e^{2.96953i}\right)
%     \left(3.17946{\times}10^{-6} e^{1.74166i}s + 0.311e^{0.10843i}\right)}.
%     \end{gather}
%     Factoring out the constant terms in the denominator yields
%     \begin{gather}
%     T.F.(x)
%     = \frac{1.0}{
%     \left(\frac{2.84723{\times}10^{-8}e^{-1.73762i}}{0.316e^{2.9695i}}x + 1\right)
%     \left(\frac{3.17946{\times}10^{-6}e^{1.74166i}}{0.311e^{0.1083i}}x + 1\right)
%     \left(0.316e^{2.9695i}\right)\left(0.311e^{0.1083i}\right)} \\
%     = \frac{1.0}{
%     \left(0.09827e^{0.9796\pi i}\right)
%     \left(\frac{e^{-1.498\pi i}}{11.098{\times}10^6}x + 1\right)
%     \left(\frac{e^{0.51\pi i}}{97.81{\times}10^3}x + 1\right)}.
%     \end{gather}

\subsection{Symbolic Regression}
A SAMPAT network was trained with 8,000 training samples of the function $f(x_1, y_1, x_2, y_2, x_3, y_3) = x_1 \cdot y_1 + x_2 \cdot y_2 + x_3 \cdot y_3$. First layer weights were constrained to be positive. The network learnt the function 
\[
\begin{split}
 0.00223903\frac{y_{1}^{0.754445}}{x_{1}^{0.120515} ~x_{2}^{0.804758} x_{3}^{0.913765} ~y_{2}^{0.902376} ~y_{3}^{0.901202}}  ~+~ 0.99943 \frac{x_{2}^{1.00032} ~x_{3}^{0.00391764} ~y_{2}^{1.00007} ~y_{3}^{0.00337599}}{x_{1}^{0.00315549} ~y_{1}^{0.00180342}} \\ + 0.998783 \frac{x_{3}^{1.00031} ~y_{2}^{0.00369535} ~y_{3}^{1.00035}}{x_{1}^{0.000617265} ~x_{2}^{0.000799159} ~y_{1}^{0.000865149}} ~+~  0.0694039\frac{x_{1}^{0.186848} ~y_{1}^{0.206967}}{x_{2}^{0.0396878} ~x_{3}^{0.203634} ~y_{2}^{0.201241} ~y_{3}^{0.177674}} \\ - 0.160931\frac{x_{1}^{0.419624} ~y_{1}^{0.686404}}{x_{2}^{0.0922101} ~x_{3}^{0.228766} ~y_{2}^{0.178757} ~y_{3}^{0.179701}} ~+~ 1.0853\frac{x_{1}^{0.925094} ~y_{1}^{0.997248}}{x_{2}^{0.0135305} ~x_{3}^{0.0328323} ~y_{2}^{0.0208514} ~y_{3}^{0.022333}}
\end{split}
\]
Terms with negligible powers ($\approx 0$) are approximately equal to 1. Neglecting terms with small weights, we note that the network approximately recovers the desired function.\\

\end{document}